\RequirePackage{iftex}

\ifLuaTeX

\fi
\documentclass[runningheads]{llncs}

\usepackage[T1]{fontenc}
\usepackage{lmodern}
\usepackage{graphicx,verbatim,multirow,xcolor,xspace,array,tikz}
\usepackage{amsmath,amssymb,amsfonts,algorithm,algpseudocode}
\usepackage{mathabx,booktabs,tabularx,xcolor,hyperref,cleveref}
\usepackage{placeins,url,orcidlink}
\usepackage[table]{xcolor}
\hypersetup{colorlinks=true,linkcolor=[rgb]{0.1,0.3,0.9},urlcolor=[rgb]{0.2,0.2,0.2},citecolor=[rgb]{0.1,0.3,0.9}}

\definecolor{palmselectgreen}{RGB}{180,230,185}
\definecolor{palmselectlightgreen}{RGB}{220,245,222}
\definecolor{secondbestgreen}{RGB}{220,245,222}
\definecolor{infeasiblered}{RGB}{255,215,215}

\begin{document}

\title{How Many Labels Are Enough? \\
ALDA: Active Learning Deployment Advisor for Medical Image Classification}
\titlerunning{ALDA}


\author{Julia Machnio \orcidlink{0009-0007-2736-2967} \and Mads Nielsen \orcidlink{0000-0003-1535-068X} \and Mostafa Mehdipour Ghazi \orcidlink{0000-0002-8473-281X}}
\authorrunning{J. Machnio et al.}
\institute{
Pioneer Centre for AI, University of Copenhagen, Copenhagen, Denmark \\
\email{\{juma,madsn,ghazi\}@di.ku.dk}
}

\maketitle

\begin{abstract} 

Active learning (AL) promises to reduce the cost of medical imaging projects by lowering the number of clinical labels required. However, practical deployment requires committing to a sampling strategy before the full annotation budget is spent, and choosing the wrong strategy can increase rather than decrease costs. We propose Active-Learning Deployment Advisor (ALDA), a deployment-oriented framework for AL method selection under clinical performance constraints. Given a short pilot phase, ALDA fits a parametric learning-curve model to each candidate strategy, estimates whether that strategy is expected to reach a required clinical performance target, and predicts the number of expert annotations needed to do so. In addition to absolute annotation cost, ALDA introduces a deployment window that quantifies the sensitivity of this cost estimate to uncertainty in the clinical threshold. The final recommendation follows a risk-aware rule: among strategies with near-optimal predicted cost, ALDA prefers the strategy with the narrowest deployment window, the most robust to threshold revisions. Experiments on four medical imaging classification domains show that ALDA predicts the deployment-optimal method from a pilot of 15--30\% of the intended budget and reduces annotation costs by up to 82\% compared with a poor strategy choice. Rather than introducing a new sampling heuristic, ALDA provides a practical decision layer that answers a deployment-critical question: how many labels are enough? 


\end{abstract}

\section{Introduction} \label{sec:intro}

Medical imaging labels are expensive because they require expert knowledge. Depending on the task, annotation may involve radiologists, pathologists, or other trained clinical specialists whose time is limited and whose expertise cannot easily be substituted. This raises a practical question: are all labels necessary?
 
Active learning (AL) addresses this question by iteratively selecting the most informative samples for annotation, aiming to reach a target performance with fewer labeled examples. In principle, AL reduces annotation cost; in practice, choosing the wrong strategy can increase it. A clinical team must select an AL method before knowing which strategy will work best for their dataset and task, yet AL method rankings are sensitive to dataset characteristics, budgets, and class-imbalance profiles \cite{hacohen2022active,machnio2026mechanism}; a method that is label-efficient in one setting may be suboptimal in another. AL deployment, therefore, becomes retrospective: annotation resources are spent first, and only afterwards does the team learn whether the chosen method was cost-effective.
 
Standard AL evaluation protocols do not address this. Methods are compared post hoc by final accuracy or area under the learning curve on fixed benchmark datasets \cite{zhan2022comparative}, answering the wrong question: not ``which method will reach our target cheapest?'' but ``which method ranked highest on these benchmarks?''. A further complication is that the clinical threshold $\tau$ is rarely a precise point value: regulatory submissions may specify a range, and the team may revise the target after initial results. A deployment framework must therefore estimate annotation cost and quantify how sensitive that estimate is to threshold uncertainty.

\begin{figure}[!b]
\centering
\includegraphics[width=0.8\linewidth]{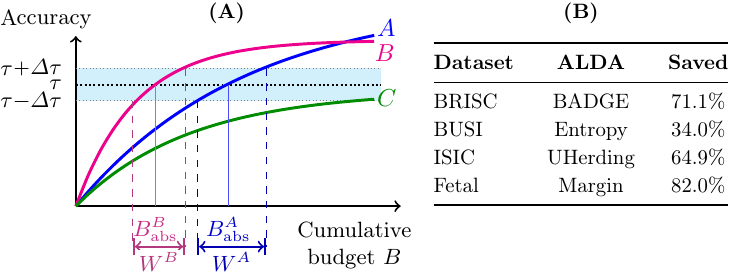}
\vspace{-0.2cm}
\caption{\textbf{(A)} Schematic of ALDA deployment quantities. The clinical target $\tau$ defines the required operating point; the shaded band denotes threshold uncertainty $\pm\Delta\tau$. Method $C$ is infeasible because its predicted ceiling lies below $\tau$. Among feasible methods, $B_{\mathrm{abs}}^B < B_{\mathrm{abs}}^A$ and $W^B < W^A$, so method $B$ is more efficient and less risky. \textbf{(B)} Labeling-cost savings of the ALDA recommendation relative to the most expensive feasible method on each dataset, predicted from a 30\% pilot (details in \Cref{tab:deployment_cost}).}
\label{fig:ALDA}
\end{figure}

Recent work on early AL curve modeling \cite{machnio2025label} has shown that AL trajectories can be forecast from partial observations using an interpretable parametric function, but trajectory forecasting alone does not yield a deployment recommendation. We adopt this parametric curve as a building block and contribute ALDA, a decision framework that turns the forecast into an actionable choice. Given a short pilot phase in which all candidate methods run for 15--30\% of the intended budget, ALDA fits the curve to each method, checks feasibility, estimates the absolute annotation cost $B_{\mathrm{abs}}(\tau)$, and computes a deployment window $W$ that quantifies sensitivity to threshold uncertainty (\Cref{fig:ALDA}). The output is a two-number deployment summary that identifies the recommended method and flags threshold-sensitive strategies before the full budget is committed.

\textit{\textbf{Our main contributions:}} (1) We formulate the AL method selection as a prospective deployment problem in medical imaging, distinct from retrospective benchmarking. (2) We introduce $B_{\mathrm{abs}}$, an absolute annotation-cost metric that answers how many labels are needed to reach a clinician-specified performance threshold, combined with a feasibility check that identifies methods predicted never to reach the target without committing the full annotation run. (3) We introduce the deployment window $W$, which measures the sensitivity of the annotation-cost estimate to uncertainty in the clinical threshold, providing an operational risk indicator for deployment planning. (4) We show that $B_{\mathrm{abs}}$ and $W$ are correlated across datasets: the cheapest feasible method is typically also among the most robust to threshold revisions, so optimizing for annotation cost simultaneously tends to minimize threshold sensitivity. (5) We show that ALDA identifies a label-efficient method from a pilot of 15--30\% of the intended budget across four medical imaging datasets. ALDA does not propose a new sampling strategy. It adds a deployment decision layer on top of existing AL methods, answering the question of whether AL is useful in practice.

\section{Related Work} \label{sec:related}
 
\paragraph{\textbf{AL sampling strategies.}}
Pool-based AL methods select unlabeled samples for annotation according to uncertainty \cite{wang2016cost}, diversity or coverage of the feature space (CoreSet \cite{sener2018active}, BADGE \cite{ash2020deep}), or geometric criteria such as typicality and probability coverage (TypiClust \cite{hacohen2022active}, ProbCover \cite{yehuda2022active}). Large-scale comparative studies \cite{hacohen2022active,machnio2026mechanism} consistently find that no single strategy dominates across tasks and budgets, providing a strong motivation for prospective method selection.
 
\paragraph{\textbf{AL in medical imaging.}}
AL has been applied across histology \cite{yang2017suggestive}, radiology \cite{tajbakhsh2020embracing}, and ultrasound classification \cite{smailagic2018medal}. Class imbalance, inter-annotator variability, and small initial pools make method rankings especially sensitive to dataset specifics, amplifying the cost of a poor choice.

\paragraph{\textbf{Learning-curve prediction.}}
Parametric learning-curve models have a long history in supervised learning \cite{mukherjee2003estimating,figueroa2012predicting}, where they support sample-size planning from small pilots. Extending such models to the AL setting is more recent: PALM \cite{machnio2025label} proposed a parametric AL curve that can be fit from partial trajectories and used to forecast final performance. We adopt this model as the curve-fitting backbone of ALDA, but contribute a separate decision layer (feasibility screening, absolute cost prediction, deployment-window analysis, and risk-aware selection), converting a trajectory forecast into a deployment recommendation.
 
\paragraph{\textbf{Stopping criteria.}}
A related but distinct line of work studies when to stop annotation given a fixed AL strategy \cite{bloodgood2009method,zhu2009active}. ALDA is complementary: it addresses which strategy to commit to before annotation begins.

\section{Methods} \label{sec:method}
 
ALDA treats AL deployment as a decision problem under uncertainty; rather than ranking methods by predicted performance alone, it estimates three quantities: feasibility, expected annotation cost, and sensitivity to the clinical target.
 
\subsection{Learning-Curve Model} \label{sec:curve}
 
We adopt the parametric AL learning curve of PALM \cite{machnio2025label}:

\begin{equation}
A = A_{\max}^{} \left[1 - \left(1-\delta^{}\right)^{\left(\frac{B}{b}+\alpha^{}\right)^{\beta^{}}}
\right] ,
\label{eq:palm}
\end{equation}

\noindent where $b > 0$ is the per-episode acquisition budget, $B$ the total annotation budget, $A_{\max} \in (0,1]$ the estimated asymptotic accuracy, $\delta \in (0,1)$ the per-episode (average coverage) gain coefficient, $\alpha \geq 0$ the budget-origin shift (seed-set contribution on early performance), and $\beta > 0$ the diminishing-returns exponent (gains scaling factor). For each method $m \in \mathcal{M}$, ALDA fits Eq. \eqref{eq:palm} to the pilot curve by nonlinear least squares (L-BFGS-B with multiple random restarts), obtaining the fit parameters $\theta_m=\{A_{\max}^{(m)},\delta^{(m)},\alpha^{(m)},\beta^{(m)}\}$.

\subsection{Clinical Target and Feasibility} \label{sec:feasibility}
 
Let $\tau$ denote the minimum acceptable performance for clinical deployment. A method is \emph{feasible} if $A_{\max}^{(m)} \geq \tau$, where its predicted ceiling reaches the clinical requirement.
Infeasible methods are excluded before any cost calculation: additional labeling is not expected to close the gap to $\tau$.

\subsection{Absolute Annotation Cost} \label{sec:cost}
 
For each feasible method, the \emph{absolute annotation cost} is the smallest budget at which predicted performance reaches $\tau$: $B_{\mathrm{abs}}^{(m)}(\tau) = \min\{B \in \mathbb{Z}_{>0} : A_m(B) \geq \tau\}$. Inverting Eq. \eqref{eq:palm} gives a closed-form expression:

\begin{equation}
B_{\mathrm{abs}}^{(m)}(\tau) = b \cdot \!\left\lceil \left(\frac{\log\bigl(1 - \tau / A_{\max}^{(m)}\bigr)}{\log\bigl(1 - \delta^{(m)}\bigr)}\right)^{\!1/\beta^{(m)}} - \alpha^{(m)} \right\rceil ,
\label{eq:babs_analytic}
\end{equation}

\noindent where $\lceil\cdot\rceil$ denotes rounding up to the nearest multiple of $\Delta_b$ (the next AL episode). Both logarithms are negative for feasible inputs ($\tau < A_{\max}$ and $\delta \in (0,1)$), so their ratio is positive, and the root is real.

\subsection{Threshold Sensitivity and Deployment Risk} \label{sec:window}
 
The clinical threshold $\tau$ may be revised after expert consultation, local validation, or regulatory review. The cheapest method at the nominal threshold is therefore not necessarily the safest: a small upward revision in $\tau$ may require many additional labels if the curve has flattened. For an accepted threshold uncertainty $\Delta\tau$ (in percentage points), the \emph{deployment risk window} is

\begin{equation}
W^{(m)} = B_{\mathrm{abs}}^{(m)}(\tau_{\mathrm{hi}}) - B_{\mathrm{abs}}^{(m)}(\tau_{\mathrm{lo}}) ,
\label{eq:window}
\end{equation}

\noindent where $\tau_{\mathrm{lo}} = \tau - \Delta\tau$ and $\tau_{\mathrm{hi}} = \tau + \Delta\tau$. To avoid evaluating $B_{\mathrm{abs}}$ arbitrarily close to the asymptote (where small errors in $A_{\max}$ produce arbitrarily large cost estimates), $\tau_{\mathrm{hi}}$ is capped per method at $A_{\max}^{(m)} - \epsilon$ for a small $\epsilon > 0$ ($\epsilon = 0.5\,\text{pp}$). A small $W^{(m)}$ indicates a stable deployment decision, whereas a large $W^{(m)}$ indicates a threshold-sensitive method; the latter is flagged as deployment-risky.

\paragraph{\textbf{Risk-aware recommendation.}}
ALDA combines expected annotation cost and deployment risk using a cost non-inferiority rule. Let
\begin{equation}
B_{\min} = \min_{m\in\mathcal{M}_{\mathrm{feas}}} B_{\mathrm{abs}}^{(m)}(\tau)
\end{equation}
be the lowest predicted annotation cost among feasible methods. Define the \emph{near-optimal cost set}
\begin{equation}
\mathcal{C}_{\eta} = \left\{m \in \mathcal{M}_{\mathrm{feas}} :
\frac{B_{\mathrm{abs}}^{(m)}(\tau)-B_{\min}}{B_{\min}} \leq \eta \right\},
\label{eq:cost_noninferior}
\end{equation}
where $\eta \geq 0$ is the tolerated relative increase in annotation
cost. The final ALDA recommendation is the lowest-risk method within this
near-optimal set:

\begin{equation}
m^* = \arg\min_{m\in\mathcal{C}_{\eta}} W^{(m)}.
\label{eq:risk_adjusted_selection}
\end{equation}

\noindent Thus, $B_{\mathrm{abs}}$ determines which methods are cost-competitive and $W$ selects the most robust option among them. The pipeline is summarized in Algorithm \ref{alg:ours}.

\begin{algorithm}[t]
\caption{ALDA: Risk-aware AL deployment}
\label{alg:ours}
\small
\begin{algorithmic}[1]
\Require Candidate AL methods $\mathcal{M}$; pilot learning curves; clinical target $\tau$; threshold uncertainty $\Delta\tau$; cost tolerance $\eta$.
\For{each method $m \in \mathcal{M}$}
\State Fit the curve $A_m$ to the pilot trajectory.
\State Estimate $\boldsymbol{\theta}_m = \{A_{\max}^{(m)},\delta^{(m)},\alpha^{(m)},\beta^{(m)}\}$.
\State Mark $m$ infeasible if $A_{\max}^{(m)} < \tau$.
\If{$m$ is feasible}
\State Compute $B_{\mathrm{abs}}^{(m)}(\tau)$ via Eq. \eqref{eq:babs_analytic}.
\State Compute $W^{(m)}$ via Eq. \eqref{eq:window}.
\EndIf
\EndFor
\State $B_{\min}\!\leftarrow\!\min_{m\in\mathcal{M}_{\mathrm{feas}}}\!B_{\mathrm{abs}}^{(m)}(\tau)$;\;
$\mathcal{C}_\eta\!=\!\{m: (B_{\mathrm{abs}}^{(m)}(\tau){-}B_{\min})/B_{\min}\!\leq\!\eta\}$.
\State $m^* = \arg\min_{m\in\mathcal{C}_{\eta}} W^{(m)}$; flag infeasible/risky methods.\\
\Return $m^{*}$, $B_{\mathrm{abs}}^{(m^{*})}$, $W^{(m^{*})}$, risk flags.
\end{algorithmic}
\end{algorithm}

\section{Results and Discussion} \label{sec:results}

\subsection{Experimental Setting} \label{sec:setup}

\paragraph{\textbf{Datasets.}}
We evaluate ALDA on four medical imaging classification datasets: BRISC2025  \cite{fateh2026brisc}, ISIC2019 \cite{gessert2020skin}, Fetal Planes \cite{burgos2020evaluation}, and BUSI \cite{al2020dataset}. Details of the utilized dataset are summarized in \Cref{tab:datasets}.

\begin{table}[ht]
\caption{Dataset summary. We report the imaging domain, input resolution, number of classes, train/test split used in our experiments, and whether the dataset is class-imbalanced. We used the test subset of ISIC2019, excluding its unknown (UNK) class.}
\label{tab:datasets}
\centering\small
\setlength{\tabcolsep}{4pt}
\resizebox{\linewidth}{!}{%
\begin{tabular}{lcccccc}
\toprule
\textbf{Dataset} & \textbf{Domain} & \textbf{Resolution} & \textbf{\#Classes} & \textbf{\#Train} & \textbf{\#Test} & \textbf{Imbalanced} \\
\midrule
BRISC2025 & Brain tumour MRI & 486$\times$475 & 4 & 5,000 & 1,000 & No \\
ISIC2019 & Skin lesion & 905$\times$936 & 8 & 4,953 & 1,238 & Yes \\
Fetal Planes & Fetal ultrasound & 548$\times$822 & 6 & 7,129 & 5,271 & Yes \\
BUSI & Breast ultrasound & 501$\times$616 & 3 & 624 & 156  & Yes \\
\bottomrule
\end{tabular}
}
\end{table}

\paragraph{\textbf{AL protocol.}}
All experiments follow a pool-based AL protocol. At each episode, an AL method selects a batch of unlabeled samples; these are added to the labeled set, and a ResNet-18 model \cite{he2016deep} is retrained on the expanded labeled set. At each AL episode, a batch of $\Delta_b = 20$ samples is acquired until exhausting the unlabeled pool, except for BUSI, where $\Delta_b = 10$. Each method--dataset pair is run for three independent random seeds; reported numbers are seed averages. Additional details are available in \Cref{app:setup}.

\paragraph{\textbf{AL strategies.}}
We compare nine AL methods spanning uncertainty-, coverage-, and representation-based selection: \textbf{Random}, \textbf{Margin}, \textbf{Entropy}, \textbf{Uncertainty}, \textbf{BADGE} \cite{ash2020deep}, \textbf{CoreSet} \cite{sener2018active}, \textbf{TypiClust} \cite{hacohen2022active}, \textbf{ProbCover}~\cite{yehuda2022active}, \textbf{UHerding} \cite{bae2025uncertainty}. ALDA is fitted to partial trajectories for each method, producing method-specific estimates of $A_{\max}$, $B_{\mathrm{abs}}(\tau)$, and $W$. Unless otherwise stated, we use a threshold uncertainty of $\Delta\tau = 5\,\text{pp}$ and a cost non-inferiority margin $\eta = 0.05$. 

\paragraph{\textbf{Evaluation.}}
We report two complementary evaluations. First, we assess whether ALDA identifies the method requiring the fewest expert labels from a 30\% pilot phase while flagging infeasible or threshold-sensitive alternatives (\Cref{sec:fullcurve}), comparing against the full AL trajectories, and report the annotations saved by the ALDA recommendation against each alternative. Second, to evaluate the prospective deployment scenario, we fit the curve to early pilot prefixes (10--30\% of the full trajectory) and measure whether the resulting ALDA recommendation matches the full-curve oracle or incurs only low label regret (\Cref{sec:pilot}).

\subsection{Full-Curve Method Selection Results} \label{sec:fullcurve}

\begin{table*}[t]
\centering
\caption{Pilot-based deployment estimates for AL method selection. ALDA is fitted to a 30\% pilot phase and used to estimate target ($\tau$) feasibility, absolute annotation cost ($B_{\mathrm{abs}}$), deployment window $W$, and the scale-free risk ratio $W/B_{\mathrm{abs}}$. $A_{ach}$ is the achieved final accuracy by the model. Dark and light greens indicate the ALDA deployment recommendations (1st and 2nd best). Red indicates methods screened as infeasible. Saved [\%] is the relative reduction in $B_{\mathrm{abs}}$ by the ALDA pick versus each alternative; negative values mark methods with marginally lower $B_{\mathrm{abs}}$ than ALDA that were rejected by the $W$ tiebreak within the $\eta$-band.}
\label{tab:deployment_cost}
\resizebox{\textwidth}{!}{%
\begin{tabular}{l ccccc ccccc}
\toprule
\multirow{2}{*}{\textbf{AL Method}}
  & \multicolumn{5}{c}{\textbf{Fetal Planes}, $\tau = 85\%$}
  & \multicolumn{5}{c}{\textbf{BRISC2025}, $\tau = 95\%$} \\
\cmidrule(lr){2-6}\cmidrule(lr){7-11}
  & $A_{\mathrm{ach}}/A_{\max} [\%]$ & $B_{\mathrm{abs}}$ & $W(m)$ & $W/B_{\mathrm{abs}}$ & Saved [\%]
  & $A_{\mathrm{ach}}/A_{\max}$ & $B_{\mathrm{abs}}$ & $W(m)$ & $W/B_{\mathrm{abs}}$ & Saved [\%]\\
\midrule
BADGE
  & 93.5 / 93.4 & 510 & 586 & 1.15 & +14.7
  & \cellcolor{palmselectgreen}99.2 / 99.2
  & \cellcolor{palmselectgreen}\textbf{895}
  & \cellcolor{palmselectgreen}\textbf{1,630}
  & \cellcolor{palmselectgreen}1.82
  & \cellcolor{palmselectgreen}\textbf{Selected} \\
CoreSet
  & 91.9 / 89.6 & 1,456 & 1,632 & 1.12 & +70.1
  & 99.2 / 99.8 & 1,351 & 3,013 & 2.23 & +33.8 \\
Entropy
  & 93.4 / 93.4 & 518 & 548 & 1.06 & +16.0
  & \cellcolor{secondbestgreen}99.2 / 99.2
  & \cellcolor{secondbestgreen}880
  & \cellcolor{secondbestgreen}1,741
  & \cellcolor{secondbestgreen}1.98
  & \cellcolor{secondbestgreen}-1.6 \\
Margin
  & \cellcolor{palmselectgreen}93.7 / 94.1
  & \cellcolor{palmselectgreen}\textbf{435}
  & \cellcolor{palmselectgreen}\textbf{590}
  & \cellcolor{palmselectgreen}1.36
  & \cellcolor{palmselectgreen}\textbf{Selected}
  & 99.2 / 100.0 & 947 & 3,330 & 3.51 & +5.6 \\
ProbCover
  & 92.5 / 99.9 & 2,418 & 1,536 & 0.64 & +82.0
  & 98.2 / 100.0 & 3,099 & 1,946 & 0.63 & +71.1 \\
Random
  & 92.1 / 93.5 & 684 & 1,156 & 1.69 & +36.4
  & 98.8 / 100.0 & 1,800 & 7,107 & 3.95 & +50.3 \\
TypiClust
  & 92.3 / 92.5 & 646 & 1,131 & 1.75 & +32.7
  & 98.4 / 100.0 & 2,603 & 11,107 & 4.27 & +65.6 \\
UHerding
  & 92.5 / 93.1 & 495 & 605 & 1.22 & +12.1
  & 99.2 / 98.9 & 933 & 1,967 & 2.11 & +4.1 \\
Uncertainty
  & \cellcolor{secondbestgreen}93.7 / 93.5
  & \cellcolor{secondbestgreen}487
  & \cellcolor{secondbestgreen}493
  & \cellcolor{secondbestgreen}1.01
  & \cellcolor{secondbestgreen}+10.7
  & 99.2 / 100.0
  & 893
  & 3,203
  & 3.59
  & -0.2 \\
\midrule\midrule
\multirow{2}{*}{\textbf{AL Method}}
  & \multicolumn{5}{c}{\textbf{ISIC2019}, $\tau = 70\%$}
  & \multicolumn{5}{c}{\textbf{BUSI}, $\tau = 75\%$} \\
\cmidrule(lr){2-6}\cmidrule(lr){7-11}
  & $A_{\mathrm{ach}}/A_{\max}$ & $B_{\mathrm{abs}}$ & $W(m)$ & $W/B_{\mathrm{abs}}$ & Saved [\%]
  & $A_{\mathrm{ach}}/A_{\max}$ & $B_{\mathrm{abs}}$ & $W(m)$ & $W/B_{\mathrm{abs}}$ & Saved [\%]\\
\midrule
BADGE
  & 73.6 / 100.0 & 4,886 & 6,633 & 1.36 & +64.9
  & 84.2 / 100.0 & 263 & 290 & 1.10 & +20.6 \\
CoreSet
  & 74.3 / 100.0 & 3,731 & 3,384 & 0.91 & +54.0
  & 82.5 / 100.0 & 316 & 270 & 0.85 & +34.0 \\
Entropy
  & \cellcolor{secondbestgreen}75.6 / 100.0
  & \cellcolor{secondbestgreen}2,195
  & \cellcolor{secondbestgreen}1,453
  & \cellcolor{secondbestgreen}0.66
  & \cellcolor{secondbestgreen}+21.9
  & \cellcolor{palmselectgreen}84.0 / 91.1
  & \cellcolor{palmselectgreen}\textbf{209}
  & \cellcolor{palmselectgreen}\textbf{257}
  & \cellcolor{palmselectgreen}1.23
  & \cellcolor{palmselectgreen}\textbf{Selected} \\
Margin
  & 75.5 / 100.0 & 2,264 & 1,949 & 0.86 & +24.2
  & \cellcolor{secondbestgreen}82.3 / 87.2
  & \cellcolor{secondbestgreen}208
  & \cellcolor{secondbestgreen}286
  & \cellcolor{secondbestgreen}1.37
  & \cellcolor{secondbestgreen}-0.5 \\
ProbCover
  & \cellcolor{infeasiblered}68.7 / 54.7
  & \cellcolor{infeasiblered}$-$
  & \cellcolor{infeasiblered}---
  & \cellcolor{infeasiblered}---
  & \cellcolor{infeasiblered}---
  & 81.4 / 78.0
  & 225
  & 208
  & 0.92
  & +7.1 \\
Random
  & 75.0 / 100.0 & 4,151 & 5,216 & 1.26 & +58.7
  & 82.7 / 85.9 & 265 & 291 & 1.10 & +21.2 \\
TypiClust
  & 74.6 / 93.1 & 3,362 & 4,056 & 1.21 & +49.0
  & 83.5 / 100.0 & 277 & 298 & 1.08 & +24.5 \\
\cellcolor{palmselectgreen}UHerding
  & \cellcolor{palmselectgreen}75.0 / 100.0
  & \cellcolor{palmselectgreen}\textbf{1,715}
  & \cellcolor{palmselectgreen}\textbf{744}
  & \cellcolor{palmselectgreen}0.43
  & \cellcolor{palmselectgreen}\textbf{Selected}
  & 83.3 / 82.1 & 233 & 261 & 1.12 & +10.3 \\
Uncertainty
  & 74.3 / 100.0 & 2,600 & 2,101 & 0.81 & +34.0
  & 83.5 / 82.1 & 233 & 227 & 0.97 & +10.5 \\
\bottomrule
\end{tabular}%
}
\end{table*}

\Cref{tab:deployment_cost} reports the annotation cost $B_{\mathrm{abs}}(\tau)$, deployment window $W$, achieved accuracy after labeling the full dataset $A_{\mathrm{ach}}$, and predicted ceiling $A_{\max}$ for every method on all datasets, computed from a 30\% pilot and compared against the full learning curves. ALDA selects the cheapest feasible method whenever a clear minimum exists, and otherwise picks the cost-competitive one with the smallest window. In feasible scenarios, absolute costs differ by up to 82\% (Margin: 435 vs.\ ProbCover: 2,418), 71\% (BADGE: 895 vs.\ ProbCover: 3,099), and 34\% (Entropy: 209 vs.\ CoreSet: 316). On BRISC2025, Entropy (880) is nominally cheapest, but BADGE (895) and Uncertainty (893) lie inside the $\eta = 5\%$ band around it; BADGE wins the $W$ tiebreak (1,630 vs.\ 1,741 for Entropy and 3,203 for Uncertainty). Representation-based methods (TypiClust, ProbCover) require 2,600--3,100 labels. Similarly, on BUSI ($\tau = 75\%$), Entropy (209, $W=257$) is preferred over the nominally cheaper Margin (208, $W=286$). ProbCover is screened as infeasible on ISIC2019 (predicted $A_{\max} = 54.7\%$, well below $\tau$).


\subsection{Pilot-Phase Reliability Results} \label{sec:pilot}

\begin{figure}[t]
\centering
\includegraphics[width=0.9\textwidth,height=0.475\textwidth]{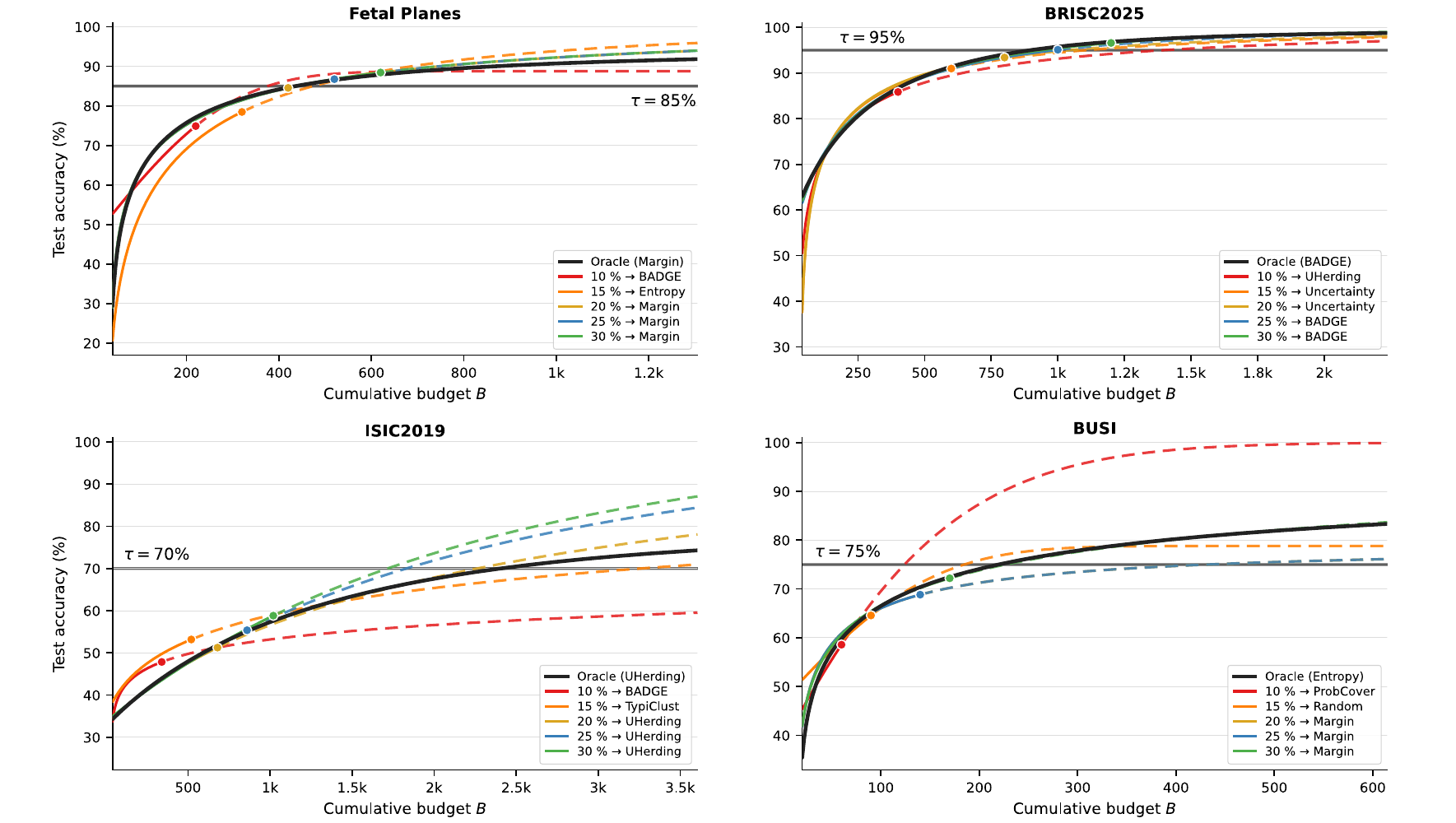}
\vspace{-0.1cm}
\caption{Pilot-based ALDA predictions from partial AL trajectories. For each dataset, the parametric curve is fitted to increasing pilot fractions (10\%, 15\%, 20\%, 25\%, 30\%) and used to extrapolate the label budget required to reach the target threshold $\tau$. The black curve denotes the oracle full-trajectory fit for the method selected on the complete trajectory; colored dashed curves show the methods selected from each pilot prefix. Dots mark the observed pilot endpoints used for fitting; dashed segments show the remaining extrapolated prediction. Horizontal lines mark the deployment target $\tau$.}
\label{fig:pilot}
\end{figure}

To evaluate the prospective deployment scenario, we fit the curve model to early prefixes of each trajectory (10--30\%) and apply ALDA. \Cref{fig:pilot} shows the method selected at each pilot fraction together with the full-curve oracle, defined as the method minimizing $B_{\mathrm{abs}}(\tau)$ alone (no $W$ tiebreak) as the strongest baseline, since it is the absolute lower bound on annotation cost. The $B_{\mathrm{abs}}$-only oracle and the risk-aware ALDA pick can differ within an $\eta$-band: on BUSI, the oracle is \emph{Margin} (208 labels) while ALDA picks \emph{Entropy} (209 labels, smaller $W$).

ALDA converges to a low-regret recommendation from small pilots. On Fetal and ISIC2019, the pilot pick matches the oracle from 20\% onward and remains stable thereafter, with predicted curves aligning with the oracle fit. On BRISC2025 and BUSI, the pick fluctuates within the same $\eta$-band before converging by 25--30\%, as predicted $B_{\mathrm{abs}}$ is not strictly monotonic in the pilot size. Sensitivity analyses for $\Delta\tau$ and $W$-cap are in \Cref{app:additional_validation}.

\section{Conclusion} \label{sec:conclusion}
 
We presented ALDA, a framework that turns AL method selection from a retrospective benchmarking exercise into a prospective deployment decision. ALDA fits a parametric learning-curve model to a short pilot, screens for feasibility, estimates the absolute cost, and quantifies sensitivity via the deployment window; the recommendation is risk-aware, among cost-competitive feasible methods, the most threshold-robust option is preferred. Across four medical imaging datasets, ALDA identifies a label-efficient method from a 15--30\% pilot and reduces annotation costs by up to 82\% relative to poor strategy choices.

ALDA's pilot phase only requires annotating a pilot subset of 20--30\% of the intended budget. Each candidate method induces an ordering over the pilot samples, yielding a method-specific partial trajectory at no extra labeling cost. ALDA fits these trajectories to estimate feasibility, target-reaching cost, and threshold sensitivity, and recommends a method on which the remaining budget is spent. Method selection thus shifts from retrospective to prospective, with the additional cost being computational (training on small pilot subsets) rather than annotation-based.


\begin{credits}

\subsubsection{\ackname} 
This project is supported by the Pioneer Centre for AI, funded by the Danish National Research Foundation (grant number P1).

\subsubsection{\discintname} 
The authors have no competing interests.

\vspace{0.5cm}\noindent\textbf{Code Availability.} 
Code is available at: \url{https://github.com/juliamachnio/PALM}.

\end{credits}

\bibliographystyle{splncs}
\bibliography{ref}

\appendix

\section{Experimental Setup and Reproducibility Details} \label{app:setup}

\paragraph{\textbf{Initialization.}}
We evaluate ALDA in a cold-start active learning (AL) setting.
For acquisition functions that require a trained model to compute scores, such as Uncertainty, Entropy, Margin, BADGE, and related uncertainty-based methods, the first acquisition batch is selected randomly. Subsequent batches are selected according to the corresponding acquisition rule.

\paragraph{\textbf{Training protocols.}}
All classifiers are trained using stochastic gradient descent (SGD) with Nesterov momentum $0.9$, weight decay $3\times10^{-4}$, and a cosine learning rate schedule with initial learning rate $0.025$. Batch size is set between 64 and 128, depending on dataset size and GPU memory. For each AL episode, the model is trained for 100 epochs on the currently labeled subset. All acquisition methods within a dataset use the same training protocol, number of epochs, evaluation split, and data preprocessing. Models are evaluated on the held-out test set after each AL episode. We use standard image augmentations, including random cropping and horizontal flipping, following common AL benchmark practice.

\paragraph{\textbf{Feature representations.}}
TypiClust and ProbCover require feature-space representations. For these methods, we extract image embeddings using a pretrained MoCo-v3 with a ResNet-50 encoder. Features are extracted from the penultimate layer and normalized using the mean and standard deviation computed from the training set. The same normalization statistics are applied to validation and test representations.

\paragraph{\textbf{Random seeds.}}
Each method--dataset pair is evaluated over three independent random seeds.
All methods use matched seeds, dataset splits, and training conditions. Unless
otherwise stated, reported values are means over seeds. Standard deviations are
reported for final test accuracy and macro-F1 in \Cref{tab:final_acc_f1}.

\paragraph{\textbf{Implementation details.}}
All experiments are implemented in PyTorch. Experiments were run on NVIDIA A100, TITAN RTX, and Quadro RTX 6000 GPUs (with 24--80 GB VRAM).

\section{Ablation Study} \label{app:additional_validation}

\subsection{Final Accuracy and Macro-F1} \label{app:acc_f1}

\Cref{tab:final_acc_f1} reports final test accuracy and macro-F1 after emptying the unlabeled pool of samples. This analysis complements the deployment-cost results in the main paper. ALDA is designed to select a method that reaches a target performance with low annotation cost and low threshold sensitivity; it is not designed to maximize final accuracy after all available labels have already been used. Nevertheless, the selected methods are generally competitive at the final episode.

Macro-F1 is particularly important for the class-imbalanced datasets, where accuracy alone can overstate performance on majority classes. Across Fetal Planes, BRISC2025, and BUSI, the relative ranking induced by final accuracy and macro-F1 is broadly consistent, with only small differences between the two metrics. ISIC2019 shows a larger gap between accuracy and macro-F1, reflecting its stronger class imbalance. Even in this setting, the ALDA-selected method remains competitive, supporting the use of the target-reaching deployment metrics in the main analysis.

\begin{table*}[t]
\centering
\small
\setlength{\tabcolsep}{4pt}
\caption{Final test accuracy and macro-F1 (mean\,$\pm$\,std over three seeds) at the last training episode for each AL method and dataset. This table complements the deployment-cost analysis by showing the final predictive performance after the full annotation budget has been consumed.}
\label{tab:final_acc_f1}
\resizebox{\textwidth}{!}{%
\begin{tabular}{l|cc|cc|cc|cc}
\toprule
\textbf{AL Method} 
& \multicolumn{2}{c|}{\textbf{Fetal Planes}} 
& \multicolumn{2}{c|}{\textbf{BRISC2025}} 
& \multicolumn{2}{c|}{\textbf{ISIC2019}} 
& \multicolumn{2}{c}{\textbf{BUSI}} \\
& Acc. & F1 
& Acc. & F1 
& Acc. & F1 
& Acc. & F1 \\
\midrule
BADGE      & $93.4{\pm}0.1$ & $92.9{\pm}0.2$ & $99.0{\pm}0.3$ & $99.1{\pm}0.1$ & $73.9{\pm}0.3$ & $60.5{\pm}0.8$ & $81.6{\pm}2.0$ & $78.7{\pm}1.1$ \\
CoreSet    & $92.2{\pm}1.1$ & $91.3{\pm}1.5$ & $99.1{\pm}0.1$ & $99.0{\pm}0.2$ & $73.3{\pm}0.9$ & $59.4{\pm}1.5$ & $81.6{\pm}2.6$ & $80.3{\pm}1.8$ \\
Entropy    & $93.1{\pm}0.2$ & $92.6{\pm}0.2$ & $99.1{\pm}0.2$ & $99.2{\pm}0.3$ & $74.1{\pm}0.3$ & $60.5{\pm}1.4$ & $82.1{\pm}2.6$ & $77.5{\pm}2.5$ \\
Margin     & $93.5{\pm}0.1$ & $92.9{\pm}0.2$ & $99.1{\pm}0.2$ & $98.9{\pm}0.2$ & $74.7{\pm}1.1$ & $60.9{\pm}2.6$ & $80.1{\pm}2.3$ & $78.0{\pm}3.9$ \\
ProbCover  & $91.9{\pm}0.2$ & $91.1{\pm}0.1$ & $97.6{\pm}0.2$ & $97.5{\pm}0.8$ & $68.0{\pm}2.5$ & $45.7{\pm}3.4$ & $79.1{\pm}2.7$ & $72.5{\pm}4.8$ \\
Random     & $93.0{\pm}1.0$ & $92.3{\pm}1.0$ & $98.8{\pm}0.4$ & $98.8{\pm}0.2$ & $75.0{\pm}1.1$ & $60.6{\pm}0.8$ & $80.1{\pm}1.7$ & $76.5{\pm}2.8$ \\
TypiClust  & $92.4{\pm}0.9$ & $91.5{\pm}1.0$ & $98.4{\pm}0.3$ & $98.5{\pm}0.7$ & $73.7{\pm}0.4$ & $60.5{\pm}3.4$ & $83.1{\pm}1.5$ & $77.6{\pm}0.6$ \\
UHerding   & $92.6{\pm}0.1$ & $91.9{\pm}0.1$ & $98.7{\pm}0.4$ & $99.0{\pm}0.0$ & $74.6{\pm}0.9$ & $62.3{\pm}1.5$ & $78.8{\pm}3.8$ & $76.1{\pm}2.8$ \\
Uncertainty& $93.3{\pm}0.1$ & $92.7{\pm}0.1$ & $99.0{\pm}0.4$ & $98.9{\pm}0.1$ & $74.5{\pm}1.1$ & $61.5{\pm}1.3$ & $79.9{\pm}1.6$ & $76.8{\pm}2.7$ \\
\bottomrule
\end{tabular}%
}
\end{table*}

\subsection{Window Half-Width Sensitivity} \label{app:window_sensitivity}

The main findings use a threshold uncertainty window of $\pm 5$ percentage points. To test whether ALDA is sensitive to this design choice, we repeat the selection procedure using several window widths.

\Cref{tab:sensitivity_delta} evaluates $\Delta\tau\in\{1,2.5,3,5,7.5,10\}$ percentage points at three target levels: the canonical threshold used in the main findings, $\tau-10$ percentage points, and $\tau-20$ percentage points. For a fixed $\tau$, $B_{\mathrm{abs}}$ is unchanged; only the deployment window $W$ changes. As expected, $W$ increases as the uncertainty interval widens. Importantly, the selected method is stable across window sizes for each fixed target level, indicating that ALDA recommendations are not an artifact of choosing $\Delta\tau=5$ percentage points.

\begin{table*}[t]
\centering
\small
\caption{Window width ($\delta$) sensitivity at three threshold levels. Each subtable fixes $\tau$ and varies $\Delta\tau$. Top: canonical $\tau$ per dataset. Middle: $\tau-10$\,pp. Bottom: $\tau-20$\,pp. For a fixed $\tau$, $B_{\mathrm{abs}}$ is constant across rows; only $W$ grows as the uncertainty window widens. \colorbox{palmselectgreen}{Green}: canonical $\delta=5$\,pp row. Consistent selections across rows and subtables indicate that ALDA recommendations are stable to window size and threshold level.}
\label{tab:sensitivity_delta}
\resizebox{\textwidth}{!}{%
\begin{tabular}{c crr crr crr crr}
\toprule
$\delta$ & \multicolumn{3}{c}{\textbf{Fetal Planes}} & \multicolumn{3}{c}{\textbf{BRISC2025}} & \multicolumn{3}{c}{\textbf{ISIC2019}} & \multicolumn{3}{c}{\textbf{BUSI}} \\
 & Selected & $B_{\mathrm{abs}}$ & $W$ & Selected & $B_{\mathrm{abs}}$ & $W$ & Selected & $B_{\mathrm{abs}}$ & $W$ & Selected & $B_{\mathrm{abs}}$ & $W$ \\
\midrule
\textit{canonical $\tau$} & \multicolumn{3}{c}{$\tau=85\%$} & \multicolumn{3}{c}{$\tau=95\%$} & \multicolumn{3}{c}{$\tau=70\%$} & \multicolumn{3}{c}{$\tau=75\%$} \\
\midrule
$\pm1.0$\,pp & Margin & 436 & 83 & BADGE & 936 & 264 & UHerding & 1{,}762 & 157 & Margin & 227 & 48 \\
$\pm2.5$\,pp & Margin & 436 & 211 & BADGE & 936 & 742 & UHerding & 1{,}762 & 394 & Margin & 227 & 121 \\
$\pm3.0$\,pp & Margin & 436 & 257 & BADGE & 936 & 957 & UHerding & 1{,}762 & 473 & Margin & 227 & 146 \\
$\pm5.0$\,pp & \cellcolor{palmselectgreen}Margin & \cellcolor{palmselectgreen}436 & \cellcolor{palmselectgreen}459 & \cellcolor{palmselectgreen}BADGE & \cellcolor{palmselectgreen}936 & \cellcolor{palmselectgreen}2{,}404 & \cellcolor{palmselectgreen}UHerding & \cellcolor{palmselectgreen}1{,}762 & \cellcolor{palmselectgreen}794 & \cellcolor{palmselectgreen}Margin & \cellcolor{palmselectgreen}227 & \cellcolor{palmselectgreen}256 \\
$\pm7.5$\,pp & Margin & 436 & 800 & BADGE & 936 & 2{,}512 & UHerding & 1{,}762 & 1{,}207 & Margin & 227 & 422 \\
$\pm10.0$\,pp & Margin & 436 & 1{,}369 & BADGE & 936 & 2{,}593 & UHerding & 1{,}762 & 1{,}642 & Margin & 227 & 645 \\
\midrule
\textit{$\tau - 10$\,pp} & \multicolumn{3}{c}{$\tau=75\%$} & \multicolumn{3}{c}{$\tau=85\%$} & \multicolumn{3}{c}{$\tau=60\%$} & \multicolumn{3}{c}{$\tau=65\%$} \\
\midrule
$\pm1.0$\,pp & Margin & 196 & 27 & Uncertainty & 305 & 51 & UHerding & 1{,}096 & 113 & Entropy & 76 & 13 \\
$\pm2.5$\,pp & Margin & 196 & 68 & Uncertainty & 305 & 130 & UHerding & 1{,}096 & 284 & Entropy & 76 & 35 \\
$\pm3.0$\,pp & Margin & 196 & 82 & Uncertainty & 305 & 158 & UHerding & 1{,}096 & 341 & Entropy & 76 & 46 \\
$\pm5.0$\,pp & \cellcolor{palmselectgreen}Margin & \cellcolor{palmselectgreen}196 & \cellcolor{palmselectgreen}141 & \cellcolor{palmselectgreen}Uncertainty & \cellcolor{palmselectgreen}305 & \cellcolor{palmselectgreen}280 & \cellcolor{palmselectgreen}UHerding & \cellcolor{palmselectgreen}1{,}096 & \cellcolor{palmselectgreen}571 & \cellcolor{palmselectgreen}Entropy & \cellcolor{palmselectgreen}76 & \cellcolor{palmselectgreen}77 \\
$\pm7.5$\,pp & Margin & 196 & 225 & Uncertainty & 305 & 479 & UHerding & 1{,}096 & 862 & Entropy & 76 & 83 \\
$\pm10.0$\,pp & Margin & 196 & 329 & Uncertainty & 305 & 799 & UHerding & 1{,}096 & 1{,}163 & Entropy & 76 & 88 \\
\midrule
\textit{$\tau - 20$\,pp} & \multicolumn{3}{c}{$\tau=65\%$} & \multicolumn{3}{c}{$\tau=75\%$} & \multicolumn{3}{c}{$\tau=50\%$} & \multicolumn{3}{c}{$\tau=55\%$} \\
\midrule
$\pm1.0$\,pp & Margin & 108 & 11 & Uncertainty & 151 & 18 & TypiClust & 350 & 90 & UHerding & 23 & 3 \\
$\pm2.5$\,pp & Margin & 108 & 28 & Uncertainty & 151 & 46 & TypiClust & 350 & 228 & UHerding & 23 & 7 \\
$\pm3.0$\,pp & Margin & 108 & 34 & Uncertainty & 151 & 55 & TypiClust & 350 & 274 & UHerding & 23 & 8 \\
$\pm5.0$\,pp & \cellcolor{palmselectgreen}Margin & \cellcolor{palmselectgreen}108 & \cellcolor{palmselectgreen}58 & \cellcolor{palmselectgreen}Uncertainty & \cellcolor{palmselectgreen}151 & \cellcolor{palmselectgreen}94 & \cellcolor{palmselectgreen}TypiClust & \cellcolor{palmselectgreen}350 & \cellcolor{palmselectgreen}469 & \cellcolor{palmselectgreen}UHerding & \cellcolor{palmselectgreen}23 & \cellcolor{palmselectgreen}17 \\
$\pm7.5$\,pp & Margin & 108 & 90 & Uncertainty & 151 & 149 & TypiClust & 350 & 737 & UHerding & 23 & 34 \\
$\pm10.0$\,pp & Margin & 108 & 127 & Uncertainty & 151 & 215 & TypiClust & 350 & 1{,}050 & UHerding & 23 & 67 \\
\bottomrule
\end{tabular}%
}
\end{table*}

\subsection{Clinical Threshold Sensitivity} \label{app:threshold_sensitivity}

We also evaluate how the ALDA recommendation changes as the clinical target $\tau$ varies. \Cref{tab:sensitivity_tau} fixes the window width to $\Delta\tau=5$ percentage points and re-runs ALDA over a range of target thresholds. This analysis addresses a practical deployment question: whether the selected AL strategy is stable under different clinical operating points.

The results show that no single AL method dominates across all thresholds and datasets. At lower thresholds, methods emphasizing representativeness or coverage (BADGE, TypiClust, Random) are often selected because early gains and broad coverage are sufficient to reach the target. At higher thresholds, the recommendation may shift toward uncertainty methods, which become more useful once the model has learned enough structure for uncertainty estimates to be informative. These transitions are expected and reflect genuine differences in the learning dynamics of AL strategies rather than the instability of ALDA.

Overall, the sensitivity analyses support the main claim of the paper: ALDA does not assume a universally best AL method. It adapts the deployment recommendation to the dataset, target threshold, and tolerated threshold uncertainty.

\begin{table*}[!t]
\centering
\small
\caption{Threshold sensitivity. ALDA is re-run at varying clinical targets $\tau$ with $\delta=5$\,pp fixed, using 30\% pilot fits. Each cell shows the selected method and annotation cost $B_{\mathrm{abs}}$. \colorbox{palmselectgreen}{Green}: canonical threshold for each dataset. Transitions between rows reveal at which target ALDA changes its deployment recommendation.}
\label{tab:sensitivity_tau}
\resizebox{\textwidth}{!}{%
\begin{tabular}{crr @{\qquad} crr @{\quad\quad} crr @{\qquad} crr}
\toprule
\multicolumn{3}{c}{\textbf{Fetal Planes}\;($\tau_0=85\%$)} & \multicolumn{3}{c}{\textbf{BRISC2025}\;($\tau_0=95\%$)} & \multicolumn{3}{c}{\textbf{ISIC2019}\;($\tau_0=70\%$)} & \multicolumn{3}{c}{\textbf{BUSI}\;($\tau_0=75\%$)} \\
$\tau$ & Selected & $B_{\mathrm{abs}}$ & $\tau$ & Selected & $B_{\mathrm{abs}}$ & $\tau$ & Selected & $B_{\mathrm{abs}}$ & $\tau$ & Selected & $B_{\mathrm{abs}}$ \\
\midrule
30\% & BADGE & 0 & 30\% & BADGE & 0 & 30\% & Entropy & 0 & 30\% & Random & 0 \\
35\% & BADGE & 0 & 35\% & BADGE & 0 & 35\% & Entropy & 0 & 35\% & Random & 0 \\
40\% & CoreSet & 0 & 40\% & BADGE & 0 & 40\% & TypiClust & 64 & 40\% & Random & 0 \\
45\% & CoreSet & 0 & 45\% & BADGE & 0 & 45\% & TypiClust & 172 & 45\% & Random & 0 \\
50\% & BADGE & 29 & 50\% & BADGE & 0 & 50\% & TypiClust & 350 & 50\% & Random & 12 \\
55\% & Random & 59 & 55\% & BADGE & 4 & 55\% & Margin & 650 & 55\% & UHerding & 23 \\
60\% & Margin & 84 & 60\% & BADGE & 28 & 60\% & UHerding & 1{,}096 & 60\% & UHerding & 36 \\
65\% & Margin & 108 & 65\% & BADGE & 59 & 65\% & UHerding & 1{,}402 & 65\% & Entropy & 76 \\
70\% & Margin & 142 & 70\% & BADGE & 100 & \cellcolor{palmselectgreen}70\% & \cellcolor{palmselectgreen}UHerding & \cellcolor{palmselectgreen}1{,}762 & 70\% & Margin & 137 \\
75\% & Margin & 196 & 75\% & Uncertainty & 151 &  &  &  & \cellcolor{palmselectgreen}75\% & \cellcolor{palmselectgreen}Margin & \cellcolor{palmselectgreen}227 \\
80\% & Margin & 284 & 80\% & Uncertainty & 208 &  &  &  &  &  &  \\
\cellcolor{palmselectgreen}85\% & \cellcolor{palmselectgreen}Margin & \cellcolor{palmselectgreen}436 & 85\% & Uncertainty & 305 &  &  &  &  &  &  \\
90\% & Uncertainty & 689 & 90\% & Uncertainty & 488 &  &  &  &  &  &  \\
 &  &  & \cellcolor{palmselectgreen}95\% & \cellcolor{palmselectgreen}BADGE & \cellcolor{palmselectgreen}936 &  &  &  &  &  &  \\
\bottomrule
\end{tabular}%
}
\end{table*}

\end{document}